\pdfoutput=1
\documentclass[11pt]{article}
\usepackage[final]{acl}
\usepackage{times}
\usepackage{latexsym}
\usepackage[T1]{fontenc}
\usepackage{cite}
\usepackage[utf8]{inputenc}
\usepackage{microtype}
\usepackage{inconsolata}
\usepackage{graphicx}
\usepackage{enumitem}
\usepackage{multirow}
\usepackage{array}
\usepackage{float}
\usepackage{hyperref}
\usepackage{multicol}
\usepackage{graphicx}
\title{The African Languages Lab: A Collaborative Approach to Advancing Low-Resource African NLP}

\author{%
\textbf{
 Sheriff Issaka$^{1}$ \quad
 Keyi Wang$^{2}$ \quad 
 Yinka Ajibola$^{3}$ \quad
 Oluwatumininu Samuel-Ipaye$^{3}$} \\
\textbf{
 Zhaoyi Zhang$^{3}$ \quad
 Nicte Aguillon Jimenez$^{3}$ \quad
 Evans Kofi Agyei$^{4}$ \quad
 Abraham Lin$^{5}$} \\
\textbf{
 Rohan Ramachandran$^{3}$ \quad
 Sadick Abdul Mumin$^{7}$ \quad
 Faith Nchifor$^{3}$ \quad
 Mohammed Shuraim$^{6}$} \\
\textbf{
 Lieqi Liu$^{1}$ \quad
 Erick Rosas Gonzalez$^{1}$ \quad
 Sylvester Kpei$^{8}$ \quad
 Jemimah Osei$^{8}$} \\
\textbf{
 Carlene Ajeneza$^{3}$ \quad
 Persis Boateng$^{9}$ \quad
 Prisca Adwoa Dufie Yeboah$^{10}$ \quad
 Saadia Gabriel$^{1}$} \\
 [1ex]
 $^{1}$University of California, Los Angeles \quad 
 $^{2}$Georgia Institute of Technology \\
 $^{3}$University of Wisconsin - Madison \quad
 $^{4}$University of Cape Coast \\
 $^{5}$Carleton University \quad
 $^{6}$Stetson University \quad
 $^{7}$Northwestern University in Qatar \\
 $^{8}$Cornell University \quad
 $^{9}$Soka University of America \quad
 $^{10}$Columbia University \\
 \texttt{\href{mailto:sheriff@cs.ucla.edu}{sheriff@cs.ucla.edu}}
}

\begin{document}
\maketitle
\begin{abstract}
Despite representing nearly one-third of the world's languages, African languages remain critically underserved by modern NLP technologies, with 88\% classified as severely underrepresented or completely ignored in computational linguistics. We present the African Languages Lab (All Lab), a comprehensive research initiative that addresses this technological gap through systematic data collection, model development, and capacity building. Our contributions include: (1) a quality-controlled data collection pipeline, yielding the largest validated African multi-modal speech and text dataset spanning 40 languages with 19 billion tokens of monolingual text and 12,628 hours of aligned speech data; (2) extensive experimental validation demonstrating that our dataset, combined with fine-tuning, achieves substantial improvements over baseline models, averaging +23.69 ChrF++, +0.33 COMET, and +15.34 BLEU points across 31 evaluated languages; and (3) a structured research program that has successfully mentored fifteen early-career researchers, establishing sustainable local capacity. Our comparative evaluation against Google Translate reveals competitive performance in several languages while identifying areas that require continued development.\footnote{To promote accessibility, we provide translations of this abstract in 10 African languages in Appendix~\ref{app:translations}, generated using our fine-tuned models.}

\end{abstract}

\section{Introduction}
The promise of artificial intelligence (AI) and natural language processing (NLP) to democratize information access remains unfulfilled for billions of speakers worldwide. Among the approximately 7,000 languages spoken globally, fewer than 20 receive substantial attention in NLP research \citep{magueresse2020lowresourcelanguagesreviewpast}. This technological marginalization particularly affects low-resource languages (LRLs). Without a clearly established definition, LRLs are languages that exist at the periphery of the digital transformation, characterized by three critical deficits: (1) a scarcity of machine-readable corpora, (2) limited personalized computational technologies and trained language models, and (3) insufficient representation in global research communities \citep{nigatu-etal-2024-zenos,issaka2024ghanaiannlplandscapelook,magueresse2020lowresourcelanguagesreviewpast}. 
While often serving substantial speaker populations, these languages face significant challenges in participating fully in the AI-driven information economy.

\begin{figure}[t]
  \includegraphics[width=\columnwidth]{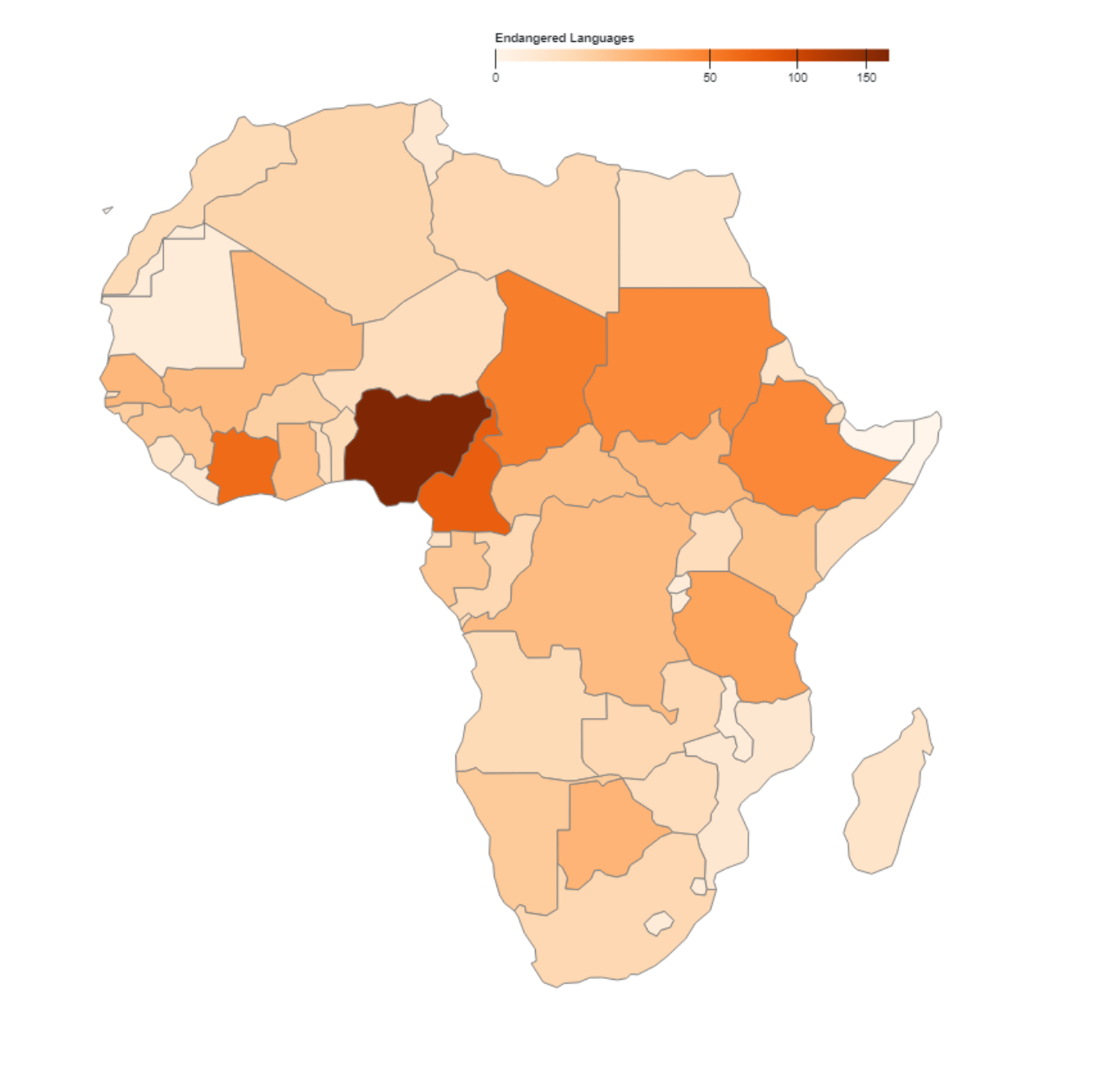}
  \caption{Number of endangered languages in each African country, where darker shading indicates a higher number of endangered languages.}
  \label{fig:language_endanger}
\end{figure}

For Africa, the scale of this crisis is staggering: over 2,000 languages are spoken across Africa (nearly one-third of all languages worldwide). Yet, a stunning 88\% of African languages are "severely underrepresented" or "completely ignored" in computational linguistics \citep{joshi-etal-2020-state}. As illustrated in Figure~\ref{fig:language_endanger}, approximately 814 African languages are in danger of extinction. Countries like Nigeria, Cameroon, and the Ivory Coast have 171, 75, and 65 languages facing the most severe threats, respectively \footnote{\url{https://www.ethnologue.com/}}. This exclusion has far-reaching consequences, from poor educational and healthcare outcomes to preventing full participation in the digital economy \citep{doi:10.1086/700617,gessler-von-der-wense-2024-nlp}.

This problem is compounded by a severe underrepresentation in the global NLP research community. Our analysis of mentions of the top 10 global languages versus the top 10 African languages across major academic databases reveals a stark imbalance. On average, for every paper discussing African languages in multilingual LLM contexts, there are 20 papers on global languages in Google Scholar (GS), 23 in COnnecting REpositories(CORE), 34 in arXiv, and 70 in The Institute of Electrical and Electronics Engineers (IEEE) (Table~\ref{tab:language-comparison} and Table~\ref{tab:language-ratios} in the Appendix). This 20-70x representation gap reinforces a self-perpetuating cycle of marginalization where limited research attention leads to poor technological support, which in turn discourages further research investment.

Contributing to broader efforts to bridge this systemic technological gap, we present the African Languages Lab (All Lab), an initiative to democratize NLP technology for African languages. Founded in 2020, the All Lab operates through a coordinated team of dedicated researchers who combine three innovative elements: 
\begin{enumerate}
    \item \textbf{Systematic data infrastructure:} We developed a systematic, quality-controlled data collection framework powered by our "All Voices" platform. All Voices is a mobile-first platform specifically designed for community-driven multilingual data collection in low-resource contexts, enabling direct translation between African languages without English intermediation
    \item \textbf{Comprehensive dataset development:} Through coordinated collection and validation efforts, we assembled the largest multi-modal dataset for African LRLs, encompassing 19 billion tokens across 40 languages with 12,628 hours of aligned speech data.
    \item \textbf{Empirical validation and capacity building:} We demonstrate the effectiveness of our approach through extensive experiments showing average improvements of +23.69 ChrF++, +0.33 COMET, and +15.34 BLEU points, while simultaneously developing local research capacity through structured mentorship programs.
\end{enumerate}

\begin{table*}[t]
  \centering
  \begin{tabular}{c|cccc||c|cccc}
    \hline
    \multicolumn{5}{c||}{\textbf{High-Resource Languages}} & \multicolumn{5}{c}{\textbf{African Languages}} \\
    \hline
    \textbf{Language} & \textbf{GS} & \textbf{arXiv} & \textbf{IEEE} & \textbf{CORE} & 
    \textbf{Language} & \textbf{GS} & \textbf{arXiv} & \textbf{IEEE} & \textbf{CORE} \\
    \hline
    \verb|English|    & 14,700 & 323 & 256 & 3,095 & \verb|Swahili|  & 617 & 10 & 3 & 114 \\
    \verb|Chinese|    & 7,710  & 60  & 85  & 1,694 & \verb|Hausa|    & 261 & 1  & 0 & 49  \\
    \verb|Hindi|      & 1,980  & 20  & 41  & 336   & \verb|Yoruba|   & 276 & 1  & 0 & 59  \\
    \verb|Spanish|    & 4,240  & 29  & 24  & 908   & \verb|Igbo|     & 203 & 0  & 0 & 38  \\
    \verb|Arabic|     & 3,150  & 25  & 24  & 616   & \verb|Amharic|  & 338 & 2  & 2 & 49  \\
    \verb|French|     & 4,490  & 38  & 17  & 1,037 & \verb|Oromo|    & 104 & 1  & 1 & 21  \\
    \verb|Bengali|    & 943    & 9   & 8   & 183   & \verb|Berber|   & 55  & 0  & 0 & 11  \\
    \verb|Portuguese| & 1,980  & 13  & 7   & 400   & \verb|Zulu|     & 175 & 1  & 1 & 38  \\
    \verb|Russian|    & 2,950  & 19  & 16  & 611   & \verb|Fula|     & 20  & 0  & 0 & 7   \\
    \verb|Urdu|       & 728    & 3   & 9   & 131   & \verb|Malagasy| & 72  & 0  & 0 & 15  \\
    \hline
  \end{tabular}
    \caption{Publication volume analysis comparing top 10 global languages versus top 10 African languages across major academic databases (2020-2024). The disparity reveals a 20-70× underrepresentation of African languages in computational linguistics research.}

  \label{tab:language-comparison}
\end{table*}

\section{Related Work}
The landscape of African NLP research has evolved through three interconnected streams: community-driven initiatives, advances in multilingual modeling, and the development of evaluation frameworks. We examine how these efforts have shaped current capabilities and identify gaps our work addresses.

\subsection{Community-Driven Research Initiatives}
The development of African NLP has been shaped by complementary community and institutional efforts. Masakhane, comprising over 3,000 Slack members, exemplifies successful community-driven research, demonstrating that participatory approaches can produce high-quality datasets and models \citep{orife2020masakhanemachinetranslation}. Complementing this work, the "Breaking the Unwritten Language Barrier" project addresses challenges specific to unwritten and under-documented languages \citep{ADDA20168}. Their work on languages like Basaa, Myene, and Embosi has established methodological approaches for speech recognition in LRLs.

These community efforts have been supported by institutional initiatives providing essential infrastructure. The Lacuna Fund has enabled dataset development \citep{rathi-etal-2023-trinity}, while Meta's No Language Left Behind project has contributed architectural innovations for massively multilingual models \citep{nllbteam2022languageleftbehindscaling}. Additional infrastructure support has come from Mozilla's Common Voice project \citep{ardila-etal-2020-common} for speech resources and the AI4D African Language Program \citep{siminyu2021ai4dafricanlanguage} for benchmark development. The Deep Learning Indaba\footnote{\url{https://deeplearningindaba.com/}} has contributed to research capacity building through its convenings, while platforms like Lanfrica have improved resource discoverability and research sharing across the continent \citep{Emezue2020LanfricaAP}.

\subsection{Advances in African Multilingual NLP}
The evolution of multilingual LLMs has shown steady progress in language coverage and capabilities. Early approaches like mBERT \citep{muller-etal-2021-unseen} and XLM-R \citep{conneau-etal-2020-unsupervised} established initial benchmarks, supporting approximately 100 languages each. Subsequent developments included more focused models like mBART \citep{liu-etal-2020-multilingual-denoising}, mT5 \citep{xue-etal-2021-mt5}, and XGLM \citep{ersoy-etal-2023-languages}, which traded broader language coverage for improved performance on specific language sets. The advent of massive LLMs further expanded these capabilities, with models like GPT-3, mGPT \citep{shliazhko-etal-2024-mgpt}, and BLOOM \citep{workshop2023bloom176bparameteropenaccessmultilingual} supporting varying numbers of African languages. Also, Glot500-m \citep{imanigooghari-etal-2023-glot500} extends support to 511 languages and the SERENGETI and Cheetah models supports about 517 African languages \citep{adebara-etal-2023-serengeti, adebara2024cheetahnaturallanguagegeneration}. Additional progress has come from the Aya model, which demonstrates instruction-following capabilities across 101 languages \citep{2024ayamodelinstructionfinetuned}, and specialized models like AfroLM, which focuses on 23 African languages \citep{dossou-etal-2022-afrolm}.

While not specifically trained in African languages, English-centric LLMs such as GPT-4 \citep{openai2024gpt4technicalreport}, Gemini \citep{geminiteam2025geminifamilyhighlycapable}, and Llama \citep{wendler-etal-2024-llamas} have shown capability in handling some African languages, \citep{robinson2023chatgptmtcompetitivehigh,ojo2024goodlargelanguagemodels,zhu2024multilingualmachinetranslationlarge,dong2024surveyincontextlearning}, though their performance generally does not match that of specialized models; underscoring the need for dedicated resources and architectures.

\subsection{Benchmarks and Evaluation Frameworks}
The development of evaluation frameworks has enabled systematic progress measurement in African NLP across diverse task domains. MasakhaNER provides NER datasets for 10 languages \citep{10.1162/tacl_a_00416}, AfriSenti offers sentiment analysis benchmarks in 14 languages \citep{muhammad2023afrisentitwittersentimentanalysis}, and AFROMT establishes standardized translation benchmarks for 8 languages \citep{reid-etal-2021-afromt}. IrokoBench unifies evaluation across natural language inference, mathematical reasoning, and multiple-choice QA in 17 African languages \citep{adelani2025irokobenchnewbenchmarkafrican}.

More targeted evaluation resources include NaijaSenti for Nigerian languages \citep{Muhammad2022NaijaSentiAN} and Kencorpus for Kenyan languages \citep{Wanjawa_Wanzare_Indede_McOnyango_Ombui_Muchemi_2023}. These Africa-focused frameworks complement broader initiatives like FLORES200 \citep{nllbteam2022languageleftbehindscaling}, the Aya Dataset \citep{singh2024ayadatasetopenaccesscollection}, and Global-MMLU \citep{singh2024globalmmluunderstandingaddressing}.

Despite these developments and advances, significant challenges remain in African NLP research \citep{adebara2022afrocentricnlpafricanlanguages, issaka2024ghanaiannlplandscapelook}. Our work builds upon these foundations while addressing several key limitations in existing approaches, such as robust team coordination, cross-initiative knowledge transfer, deduplication of efforts, and intentional skill set development.

\section{Methodology}

\subsection{Datasets}
\noindent\textbf{All Voices Platform.} To address the fundamental challenge of data scarcity in African languages, we developed All Voices, a mobile-first platform that stands as the only solution specifically designed for data collection in any LRL. The platform’s innovative approach enables direct translation between LRLs without requiring English as an intermediary, addressing a critical gap in the existing data collection infrastructure. In addition, All Voices distinguishes itself through its multimodal capabilities, which support the collection and validation of text and audio data. The platform features an intuitive, user-friendly interface that encourages broad participation, complemented by gamification elements, including a global leaderboard system that promotes user engagement. Importantly, All Voices is open and free to everyone, aligning with our mission to democratize language technology development. All Voices contributors provide informed consent for their contributions to be used for research purposes, including dataset creation, model training, and open-source distribution, with full transparency regarding data usage and the right to withdraw consent at any time.

\begin{figure}[t]
  \includegraphics[width=\columnwidth]{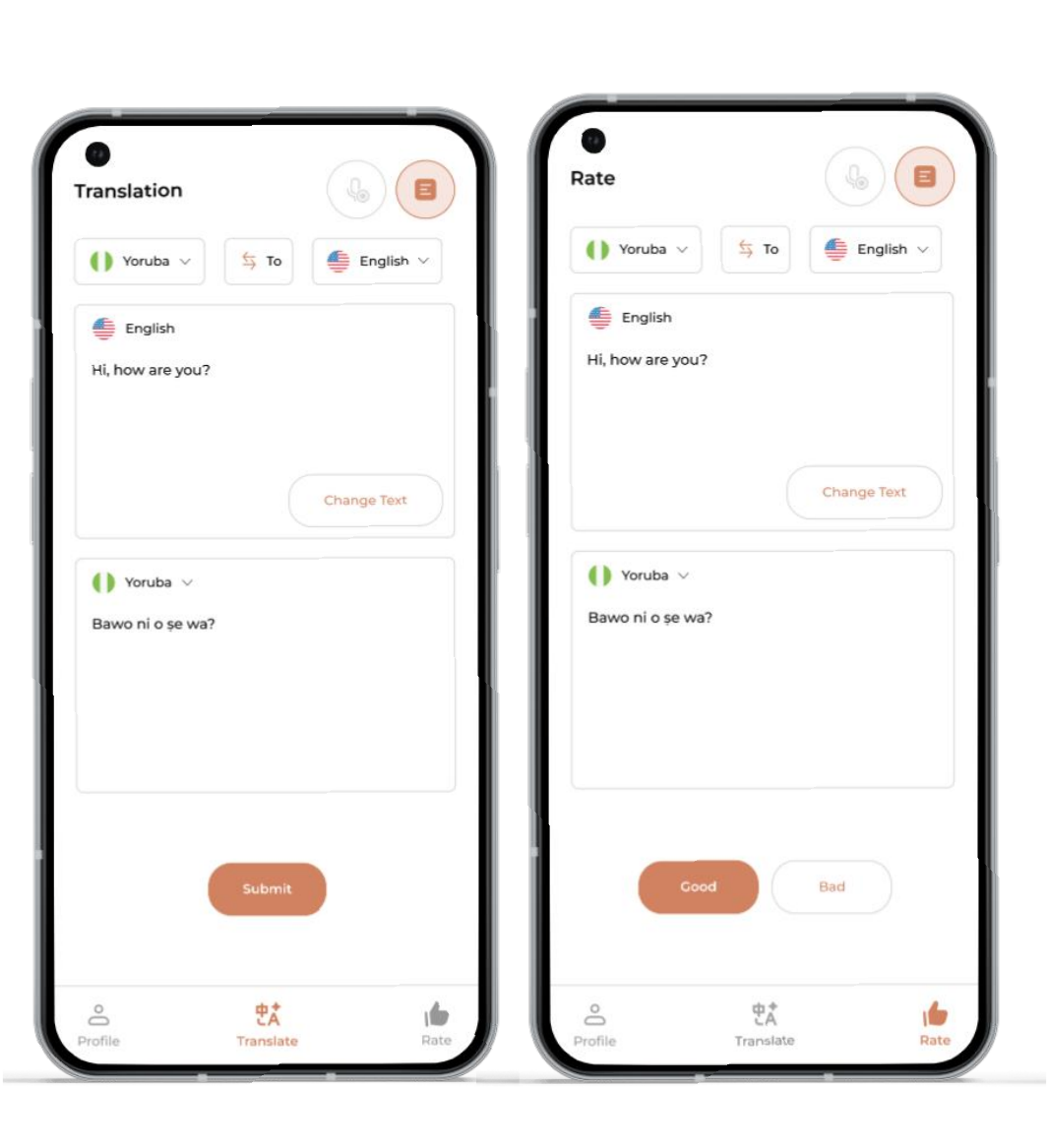}
  \caption{The All Voices platform interface demonstrating its dual functionality: direct text translation from English to Yoruba (left panel) and community-driven translation validation system (right panel). The mobile-first design enables participation from users with limited technical resources.}
  \label{fig:all_voices}
\end{figure}

The platform’s architecture, built using ReactNative~\footnote{\url{https://reactnative.dev/}} and Firebase~\footnote{\url{https://firebase.google.com/}}, integrates user authentication and analytics, translation corpus management, and quality control components. Our authentication system provides comprehensive user profiling, tracking contributor demographics and expertise through quantifiable metrics, including successful translations and community validation scores. This system implements OAuth 2.0 authentication and role-based access control to ensure data integrity and user privacy. The translation corpus management system centrally stores both text and audio translations along with their metadata, and protects all data using AES-256 encryption at rest and TLS 1.3 during transmission. Translations undergo peer review requiring both a minimum threshold of positive validation (\textgreater5 upvotes) and an acceptable error margin (\textless3 downvotes) to achieve verified status. A key innovation is our recursive translation pipeline: verified translations become eligible source material for subsequent translations, creating a multiplicative effect in data collection.

\noindent\textbf{Data Collection and Processing.} Our dataset development methodology combines crowd-sourced translations through All Voices (Figure~\ref{fig:all_voices}) with carefully curated open-source corpora. We integrate validated translations from our platform with established datasets, including NLLB \citep{nllbteam2022languageleftbehindscaling}, CCMatrix \citep{wenzek2019ccnetextractinghighquality}, OpenSubtitles \citep{tiedemann-2016-finding}, MultiCCAligned \citep{elkishky_ccaligned_2020}, ParaCrawl \citep{banon-etal-2020-paracrawl}, XLEnt \citep{el-kishky-etal-2021-xlent}, MultiParaCrawl\citep{banon-etal-2020-paracrawl}, LinguaTools-WikiTitles \citep{TIEDEMANN12.463}, and CCAligned \citep{elkishky_ccaligned_2020}. Additionally, we collect new datasets through our community partners.

\begin{figure}
    \centering
    \includegraphics[width=1\columnwidth]{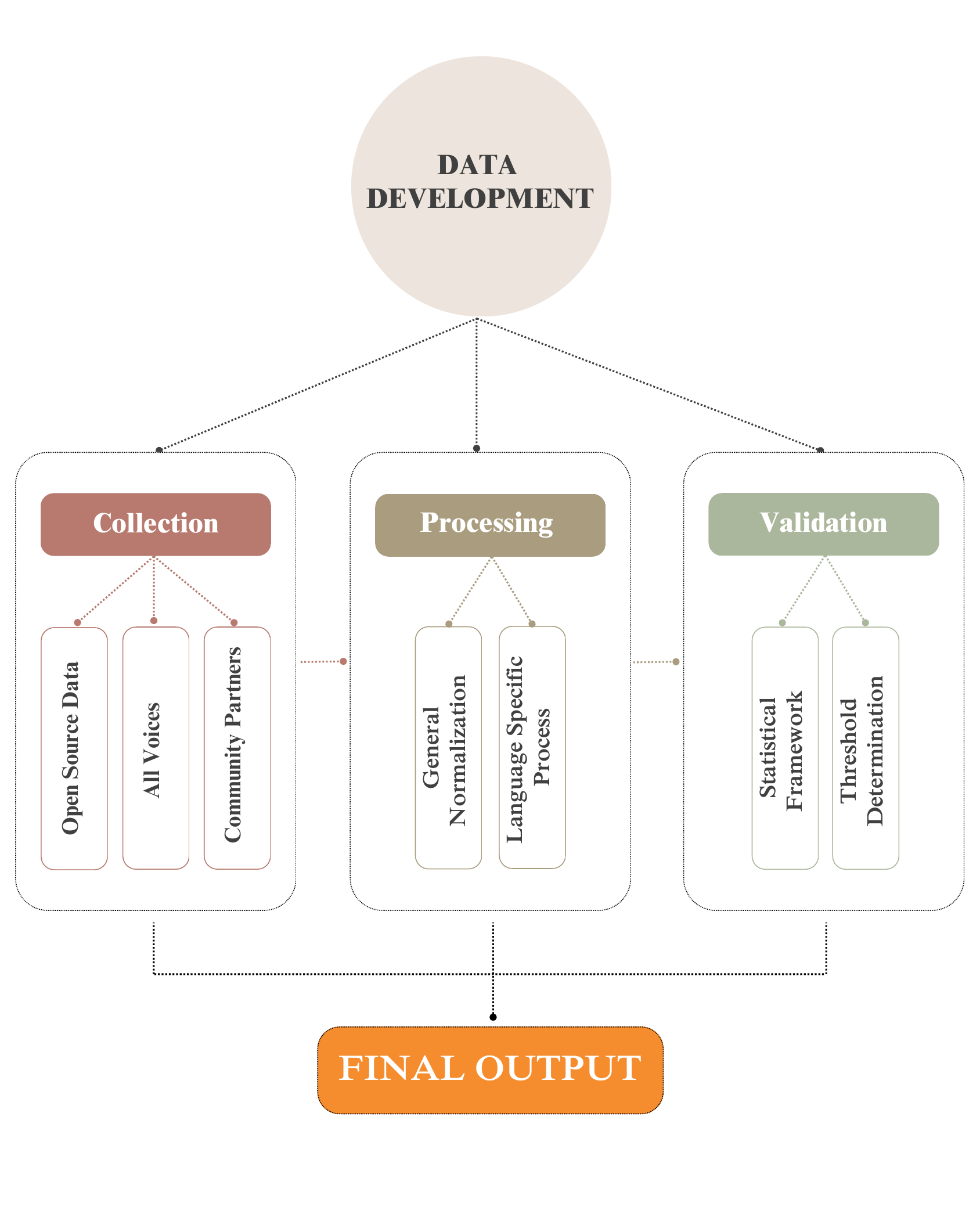}
    \caption{End-to-end processing pipeline showing multi-source integration, language-specific preprocessing, and statistical validation ensuring dataset quality.}
    \label{fig:pipeline}
\end{figure}

Our data processing implements a robust two-tier approach combining general normalization with language-specific processing. The general normalization phase addresses universal text artifacts through Unicode normalization, character encoding standardization, and structural cleaning, including HTML removal and symbol standardization. The language-specific processing phase implements specialized handling for African language features, including morphological analysis, script variant normalization, and tone mark standardization, language identification, with custom rule sets developed for specific language families.

Next, our translation validation methodology implements a robust statistical framework for assessing translation quality through quantitative analysis of character-level distributions. The validation metric employs character ratio analysis between source and target texts, computed as the ratio of target text length to source text length. We analyze these ratios using z-score normalization within language-specific distribution, enabling the detection of statistical outliers while accounting for natural variations in text length across different language pairs. This approach is augmented with character overlap detection to identify potential artifacts or inappropriate text preservation, particularly crucial for languages sharing similar orthographic features.

Also, the threshold determination process implements an adaptive sampling methodology. For each language pair, we establish baseline distributions through initial sampling of 10,000 translation pairs, employing Kernel Density Estimation for robust distribution modeling. This approach effectively captures the non-Gaussian characteristics frequently observed in cross-lingual character distributions. Thresholds are dynamically computed using a modified Tukey method with an adaptive multiplier. This adaptive threshold mechanism automatically calibrates to language-specific characteristics, implementing more stringent filtering for language pairs that exhibit consistent ratios while allowing appropriate flexibility for pairs with inherently higher variability. The resulting validation framework effectively identifies and filters anomalous translations while maintaining sensitivity to legitimate linguistic variations across diverse African language families. The processed data sets are structured according to HuggingFace~\footnote{\url{https://huggingface.co/}} Dataset specifications, enabling seamless API integration.

\subsection{Model Development}
To evaluate the utility of our dataset and establish baselines for our languages, we experimented with Llama-3.2-1B \citep{grattafiori2024llama3herdmodels}. We chose this model as our base because of its demonstrated multilingual capabilities and efficient parameter scaling, making it suitable for LRLs. Critically, while capable of tokenizing, this model has not been explicitly trained on any African language in our dataset. Thus, providing a cleaner baseline for measuring the utility of our dataset during fine-tuning without risk of data contamination. 

We employed full fine-tuning rather than parameter-efficient methods, as preliminary experiments with Quantization-aware Low-Rank Adaptation (QLoRA) \citep{2024ayamodelinstructionfinetuned} yielded insufficient performance gains for our target languages. Our training pipeline uses supervised learning with a standardized instruction template: "Translate the following English text to X:", where X denotes the target African language.

Training leveraged NVIDIA H100 GPUs with the following parameters: batch size of 64 with 4-step gradient accumulation, maximum sequence length of 1024 tokens for both input and output, single epoch training using all available parallel data per language, learning rate of $5.0 \times 10^{-5}$ with cosine scheduling and 0.15 warm-up ratio, and BF16 mixed precision for memory efficiency. During inference, we maintained consistency with batch size 64 while using temperature 0.1, top-p 0.95, top-k 50, and maximum output length of 1024 tokens to balance diversity with quality and reproducibility.

\subsection{Evaluation Metrics}
Model performance was evaluated using a complementary set of metrics: BiLingual Evaluation Understudy (BLEU) \citep{wieting-etal-2019-beyond}, which measures n-gram precision; METEOR \citep{banerjee-lavie-2005-meteor}, which accounts for word stems and synonyms; COMET \citep{rei2020cometneuralframeworkmt}, which leverages multilingual embeddings to assess semantic similarity; and ChrF++ \citep{wang2025llmsreplacehumanevaluators}, which operates on character-level n-grams to better capture morphological variations common in African languages. Additionally, we employed Translation Edit Rate (TER) \citep{snover-etal-2006-study}, which quantifies the minimum number of edits required to transform the hypothesis into the reference translation (where lower scores indicate better quality), and AfriCOMET \citep{wang2024afrimteafricometenhancingcomet}, a neural metric specifically trained on African language pairs to better capture language-specific quality nuances.

Together, these metrics comprehensively assess translation quality across different linguistic aspects, from surface-level n-gram matching to semantic preservation and post-editing effort. We utilized the FLORES-200 dataset \citep{nllbteam2022languageleftbehindscaling} as our standardized test set, ensuring consistency across all languages and enabling direct comparison with other multilingual systems.

\section{Results}

\begin{table*}[t]
  \centering
  \begin{tabular}{lcc|lcc|lcc}
    \hline
    \textbf{Language} & \textbf{Tokens} & \textbf{Hours} & \textbf{Language} & \textbf{Tokens} & \textbf{Hours} & \textbf{Language} & \textbf{Tokens} & \textbf{Hours} \\
    \hline
    Amharic & 2,944.95 & 238.00 & Sesotho & 274.61 & 114.70 & Tshiluba & 54.93 & - \\
    Arabic & 2,400.00 & 2,721.52 & Oromo & 252.82 & 145.00 & Mossi & 50.59 & - \\
    Yoruba & 2,362.70 & 128.30 & Chewa & 230.63 & 35.00 & Kikongo & 46.59 & - \\
    Afrikaans & 2,295.09 & 138.00 & Rundi & 172.61 & - & Ewe & 31.74 & 147.00 \\
    Hausa & 1,538.84 & 239.00 & Luganda & 121.17 & 1,727.80 & Berber & 28.86 & 19.33 \\
    Tigrinya & 916.42 & 1.00 & Tswana & 118.84 & 111.70 & Krio & 22.76 & 80.00 \\
    Malagasy & 839.12 & 325.14 & Bambara & 109.49 & 30.60 & Bemba & 8.60 & 230.30 \\
    Somali & 751.13 & 115.40 & Lingala & 102.19 & 194.30 & Kanuri & 6.18 & - \\
    Swahili & 700.39 & 1,115.00 & Twi & 86.49 & 227.03 & Umbundu & 5.10 & - \\
    Xhosa & 563.07 & 123.70 & Fon & 77.27 & 18.50 & Kiluba & 2.02 & - \\
    Zulu & 553.67 & 83.20 & Fula & 72.40 & 124.00 & Ngambay & 1.03 & - \\
    Igbo & 433.28 & 25.00 & Kikuyu & 66.34 & 44.00 & Mandinka & 0.41 & - \\
    Shona & 428.25 & 103.00 & Wolof & 57.46 & 183.20 & Fang & 0.02 & - \\
    Kinyarwanda & 283.40 & 3,839.00 & & & & & & \\
    \hline
  \end{tabular}
  \caption{Dataset composition across our 40 African languages sorted by token count, showing the distribution of tokens (in millions) and hours of audio data. Dashes (-) indicate no audio data available.}
  \label{tab:merged_dataset}
\end{table*}

\begin{table*}[t]
  \centering
  \resizebox{\textwidth}{!}{%
  \begin{tabular}{c|ccc|ccc|ccc|ccc|ccc|ccc}
    \hline
    \multirow{2}{*}{\textbf{Language}} 
    & \multicolumn{3}{c|}{\textbf{ChrF++}~($\uparrow$)} 
    & \multicolumn{3}{c|}{\textbf{COMET}~($\uparrow$)} 
    & \multicolumn{3}{c|}{\textbf{BLEU}~($\uparrow$)} 
    & \multicolumn{3}{c|}{\textbf{Africomet}~($\uparrow$)} 
    & \multicolumn{3}{c|}{\textbf{METEOR}~($\uparrow$)} 
    & \multicolumn{3}{c}{\textbf{TER}~($\downarrow$)} \\
    & \textbf{Llama1B} & \textbf{Ours} & \textbf{GT} 
    & \textbf{Llama1B} & \textbf{Ours} & \textbf{GT} 
    & \textbf{Llama1B} & \textbf{Ours} & \textbf{GT} 
    & \textbf{Llama1B} & \textbf{Ours} & \textbf{GT} 
    & \textbf{Llama1B} & \textbf{Ours} & \textbf{GT} 
    & \textbf{Llama1B} & \textbf{Ours} & \textbf{GT} \\
    \hline

    \texttt{Amharic} & 3.26 & 28.69 & 30.24 & 0.38 & 0.82 & 0.88 & 3.83 & 12.16 & 16.11 & 0.14 & 0.56 & 0.72 & 0.01 & 0.30 & 0.40 & 58.40 & 64.24 & 58.40 \\
    \texttt{Fula} & 2.97 & 18.73 & - & 0.32 & 0.55 & - & 0.14 & 5.73 & - & 0.03 & 0.15 & - & 0.08 & 0.06 & - & 1151.08 & 72.22 & - \\
    \texttt{Yoruba} & 3.77 & 30.88 & 21.05 & 0.23 & 0.67 & 0.56 & 0.41 & 32.33 & 11.96 & 0.01 & 0.60 & 0.55 & 0.05 & 0.27 & 0.16 & 525.95 & 84.31 & 76.28 \\
    \texttt{Igbo} & 5.30 & 33.42 & 45.92 & 0.27 & 0.71 & 0.72 & 0.65 & 14.10 & 36.09 & -0.70 & 0.52 & 0.57 & 0.06 & 0.41 & 0.43 & 523.93 & 73.11 & 52.80 \\
    \texttt{Oromo} & 11.30 & 27.74 & 54.93 & 0.31 & 0.70 & 0.80 & 5.54 & 5.72 & 33.57 & 0.13 & 0.29 & 0.66 & 0.06 & 0.12 & 0.29 & 77.62 & 119.01 & 56.92 \\
    \texttt{Swahili} & 9.0 & 72.27 & 75.81 & 0.40 & 0.78 & 0.85 & 0.54 & 56.23 & 54.81 & -0.20 & 0.62 & 0.72 & 0.07 & 0.57 & 0.65 & 1021.93 & 23.77 & 23.77 \\
    \texttt{Hausa} & 7.75 & 51.77 & 54.60 & 0.39 & 0.70 & 0.80 & 0.57 & 22.38 & 45.49 & 0.01 & 0.47 & 0.64 & 0.07 & 0.41 & 0.53 & 993.66 & 52.3 & 64.37 \\
    \texttt{Twi(Asante)} & 4.0 & 46.80 & 31.48 & 0.26 & 0.71 & 0.71 & 0.29 & 27.36 & 17.81 & -0.06 & 0.30 & 0.36 & 0.07 & 0.29 & 0.33 & 656.31 & 49.12 & 64.24 \\
    \texttt{Shona} & 7.52 & 36.55 & 51.93 & 0.27 & 0.60 & 0.63 & 0.25 & 12.06 & 18.05 & 0.01 & 0.47 & 0.59 & 0.07 & 0.28 & 0.34 & 1318.15 & 99.15 & 64.16 \\
    \texttt{Kinyarwanda} & 5.62 & 24.65 & 70.27 & 0.28 & 0.56 & 0.67 & 0.37 & 17.25 & 49.60 & -0.18 & 0.40 & 0.66 & 0.04 & 0.23 & 0.48 & 428.45 & 65.53 & 30.24 \\
    \texttt{Ewe} & 3.05 & 33.48 & 47.86 & 0.22 & 0.26 & 0.37 & 0.27 & 31.55 & 25.79 & 0.06 & 0.24 & 0.33 & 0.07 & 0.26 & 0.37 & 689.73 & 93.32 & 48.69 \\
    \texttt{Bambara} & 4.06 & 15.60 & 27.02 & 0.25 & 0.65 & 0.72 & 0.18 & 22.50 & 13.60 & 0.08 & 0.22 & 0.39 & 0.08 & 0.20 & 0.29 & 766.87 & 49.33 & 112.12 \\
    \texttt{Wolof} & 2.00 & 12.01 & - & 0.30 & 0.61 & - & 0.16 & 2.68 & - & 0.07 & 0.25 & - & 0.07 & 0.28 & - & 1030.14 & 84.83 & - \\
    \texttt{Luganda} & 7.08 & 24.91 & 23.55 & 0.28 & 0.64 & 0.63 & 0.41 & 3.84 & 3.96 & -0.05 & 0.55 & 0.57 & 0.07 & 0.31 & 0.33 & 980.96 & 74.57 & 63.10 \\
    \texttt{Arabic} & 8.67 & 31.52 & 28.46 & 0.52 & 0.85 & 0.89 & 0.30 & 23.36 & 21.41 & 0.21 & 0.70 & 0.77 & 0.08 & 0.52 & 0.55 & 1164.70 & 67.00 & 67.00 \\
    \texttt{Somali} & 7.11 & 35.64 & 47.32 & 0.37 & 0.76 & 0.80 & 0.41 & 8.99 & 12.90 & 0.03 & 0.51 & 0.62 & 0.07 & 0.32 & 0.37 & 442.60 & 65.09 & 60.75 \\
    \texttt{Afrkaans} & 44.76 & 48.07 & 52.10 & 0.68 & 0.86 & 0.85 & 32.98 & 18.00 & 27.25 & 0.37 & 0.74 & 0.73 & 0.35 & 0.66 & 0.65 & 52.61 & 48.23 & 39.46 \\
    \texttt{Tigrinya} & 5.43 & 24.12 & 23.75 & 0.27 & 0.76 & 0.83 & 2.77 & 13.65 & 13.00 & 0.04 & 0.43 & 0.70 & 0.01 & 0.16 & 0.23 & 78.33 & 36.55 & 78.33 \\
    \texttt{Malagasy} & 9.39 & 26.29 &  43.06 & 0.42 & 0.80 & 0.76 & 0.32 & 22.49 &  9.10 & 0.13 & 0.61 & 0.45 & 0.06 & 0.39 & 0.36 & 856.18 & 57.08 & 57.08 \\
    \texttt{Xhosa} & 6.58 & 46.17 & 47.88 & 0.35 & 0.75 & 0.70 & 0.63 & 12.87 & 14.95 & 0.21 & 0.55 & 0.41 & 0.09 & 0.33 & 0.31 & 561.279 & 71.79 & 58.74 \\
    \texttt{Zulu} & 7.58 & 38.06 & 64.45 & 0.32 & 0.76 & 0.73 & 0.37 & 5.90 & 34.57 & 0.09 & 0.62 & 0.54 & 0.07 & 0.41 & 0.35 & 728.98 & 89.52 & 51.16 \\
    \texttt{Sesotho} & 7.09 & 61.79 &  57.01 & 0.28 & 0.61 & 0.66 & 0.14 & 51.16 & 33.25 & 0.14 & 0.59 & 0.65 & 0.08 & 0.31 & 0.44 & 1475.22 & 63.22 & 45.66 \\
    \texttt{Chewa} & 4.76 & 28.57 & 34.29 & 0.28 & 0.62 & 0.61 & 0.0 & 13.00 & 12.67 & -0.02 & 0.56 & 0.44 & 0.07 & 0.35 & 0.31 & 876.28 & 72.16 & 67.01 \\
    \texttt{Lingala} & 5.91 & 33.32 & 38.14 & 0.30 & 0.58 & 0.65 & 0.23 & 18.21 & 16.82 & 0.00 & 0.12 & 0.40 & 0.08 & 0.23 & 0.45 & 955.39 & 80.32 & 54.96 \\
    \texttt{Tswana} & 8.59 & 26.37 & - & 0.28 & 0.63 & - & 0.69 & 16.98 & - & 0.11 & 0.38 & - & 0.07 & 0.27 & - & 413.87 & 65.69 & - \\
    \texttt{Fon} & 10.57 & 9.47 & - & 0.16 & 0.60 & - & 5.14 & 6.50 & - & 0.14 & 0.04 & - & 0.06 & 0.06 & - & 56.15 & 29.73 & - \\
    \texttt{Kikuyu} & 9.77 & 25.80 & - & 0.27 & 0.55 & - & 0.28 & 17.92 & - & 0.03 & 0.03 & - & 0.07 & 0.09 & - & 943.70 & 59.38 & - \\
    \texttt{Tshiluba} & 15.68 & 22.10 & - & 0.31 & 0.55 & - & 5.856 & 7.44 & - & 0.20 & 0.12 & - & 0.11 & 0.11 & - & 61.49 & 52.71 & - \\
    \texttt{Mossi} & 12.82 & 23.99 & - & 0.28 & 0.51 & - & 5.86 & 9.62 & - & 0.18 & -0.03 & - & 0.08 & 0.09 & - & 57.02 & 81.46 & - \\
    \texttt{Kikongo} & 6.18 & 20.84 & - & 0.31 & 0.55 & - & 0.38 & 6.15 & - & -0.05 & 0.07 & - & 0.05 & 0.16 & - & 318.97 & 44.51 & - \\
    \texttt{Bemba} & 3.59 & 25.79 & - & 0.31 & 0.46 & - & 0.25 & 27.69 & - & 0.09 & 0.07 & - & 0.10 & 0.10 & - & 755.92 & 48.63 & - \\
    \texttt{Average} & 8.10 & 31.79 & 44.14 & 0.32 & 0.65 & 0.72 & 2.27 & 17.61 & 23.76 & 0.04 & 0.38 & 0.57 & 0.08 & 0.28 & 0.39 & 645.87 & 65.74 & 58.87 \\
    \hline
  \end{tabular}%
  }
  \caption{Performance comparison between base Llama-3.2-1B (Llama1B), our Finetuned Llama-3.2-1B models (Ours), and Google Translate (GT) across different metrics (ChrF++, COMET, BLEU, Africomet, METEOR, and TER). Higher scores indicate better performance (except for TER, where lower scores are better).}
  \label{tab:model-comparison}
\end{table*}

Our comprehensive data collection yielded 19,012,019,696 tokens of monolingual text and 12,628 hours of aligned audio across 40 African languages (Table~\ref{tab:merged_dataset}). Our analysis reveals distinct stratification patterns highlighting the digital divide within African languages themselves.

From a text perspective (measured in millions of tokens), we observe four distinct tiers:
\begin{enumerate}
    \item \textbf{Primary resource languages (>2B tokens):} This tier includes Amharic (2.94B), Arabic (2.40B), Yoruba (2.36B), and Afrikaans (2.30B), reflecting sustained digitization efforts and strong institutional support.
    \item \textbf{Established digital languages ($\sim$1–2B tokens):} Languages such as Hausa (1.54B) and Tigrinya (0.92B) demonstrate robust digital presence, likely owing to consistent documentation and preservation initiatives.
    \item \textbf{Emerging digital languages (250M–1B tokens):} A substantial group including Malagasy (839.12M), Somali (751.13M), Swahili (700.39M), and Xhosa (563.07M), show growing digital footprints but still lagging behind the top tiers.
    \item \textbf{Resource-constrained languages (\textless250M tokens):} The majority of languages in our dataset fall into this category, including widely spoken languages such as Bambara (109.49M) and Luganda (121.17M). This tier reflects substantial gaps in textual data availability.
\end{enumerate}

For audio resources (measured in hours), the stratification follows a different pattern, highlighting a distinct set of leading languages:

\begin{enumerate}
    \item \textbf{High-resource audio languages (\textgreater1,000 hours):} Kinyarwanda (3,839.00h), Luganda (1,727.80h), Swahili (1,115.00h), and Arabic (2,721.52h) dominate in audio availability, often due to large-scale speech corpora or broadcast archives.
    \item \textbf{Established audio languages (500–1,000 hours):} This tier is notably sparse, underscoring the scarcity of mid-scale speech datasets in African languages.
    \item \textbf{Moderate audio languages (100–500 hours):} Includes Malagasy (325.14h), Twi (227.03h), Bemba (230.30h), and Ewe (147.00h), representing a mix of widely spoken languages and those with targeted speech collection efforts.
    \item \textbf{Low-resource audio languages (<100 hours):} Many languages, including Kikongo, Rundi, Kanuri, Umbundu, and Fang, have either minimal or no audio data.
\end{enumerate}

Overall, our analysis underscores two parallel digital divides: a textual divide, where a small set of languages capture the majority of tokens, and an audio divide, where a different but equally narrow set of languages dominate. Notably, the top three languages by text volume account for a disproportionate share of tokens, while the top three by audio hours similarly capture the bulk of recorded speech. This imbalance highlights the urgent need for targeted development of both textual and audio resources, particularly for languages with substantial speaker populations but limited digital presence.

\subsection{Translation Performance Analysis}
Our experimental evaluation across 31 African languages reveals substantial and systematic improvements through fine-tuning, with distinct patterns emerging across language families and resource levels (Table~\ref{tab:model-comparison}). Nine languages were excluded from evaluation due to insufficient training data or absence from the FLORES-200 benchmark.

\noindent\textbf{Baseline Performance.} The base Llama-3.2-1B model demonstrates limited but non-trivial capability for African languages, revealing interesting patterns of cross-lingual transfer. ChrF++ scores range from 2.00 (Wolof) to 44.76 (Afrikaans), with a mean of 8.10, indicating minimal character-level understanding for most languages. COMET scores cluster between 0.16-0.68 (mean: 0.32), suggesting some semantic comprehension despite poor surface realization. Notably, Afrikaans shows exceptional baseline performance (ChrF++ 44.76, BLEU 32.98), leveraging its Germanic roots and Latin script. The extremely low baseline BLEU scores (mean: 2.27) across most languages confirm the model's inability to produce accurate n-gram sequences without language-specific training.

\noindent\textbf{Fine-tuning Impact.} Our dataset enables noticeable performance improvements, with average gains of +23.69 ChrF++, +0.33 COMET, and +15.34 BLEU points. The magnitude of improvement correlates inversely with baseline performance, suggesting effective transfer learning rather than simple memorization. Swahili exhibits the largest absolute ChrF++ improvement (+63.27 points), achieving near-parity with Google Translate (72.27 vs 75.81). Sesotho shows remarkable gains across all metrics (+61.79 ChrF++, +51.16 BLEU), while maintaining competitive performance against Google Translate. Languages with minimal baselines, including Fula, Wolof, and Kikongo, demonstrate that even severely under-resourced languages benefit substantially from targeted fine-tuning, achieving functional translation capability where none existed before.

\noindent\textbf{Comparison with Google Translate.} Where available (22 languages), comparison with Google Translate reveals three distinct performance categories. First, languages where our models achieve competitive or superior performance: Yoruba (30.88 vs 21.05 ChrF++), Arabic (31.52 vs 28.46), and notably Twi, where we substantially outperform Google Translate (46.80 vs 31.48). Second, languages where Google maintains clear advantages, particularly in high-resource cases like Kinyarwanda (24.65 vs 70.27) and Swahili (72.27 vs 75.81), reflecting their extensive training data. Third, the languages where Google Translate offers no support, highlighting our contribution to genuinely LRL coverage.

\noindent\textbf{Language-Specific Patterns.} Three response profiles emerge from our analysis. High-responder languages (Swahili, Sesotho, Hausa) show dramatic improvements exceeding 40 ChrF++ points, suggesting optimal alignment between our dataset characteristics and model architecture. Steady improvers (Igbo, Shona, Somali) demonstrate consistent gains of 25-30 points across metrics, indicating robust but not exceptional adaptation. Challenging cases (Fon, Wolof, Bambara) show limited improvements despite fine-tuning, likely requiring specialized tokenization or architectural modifications to address their unique linguistic features.

\noindent\textbf{Translation Edit Rate Analysis.} The dramatic TER reductions, averaging 580.13 points lower after fine-tuning, provide crucial practical insights. Languages like Swahili achieve TER scores comparable to Google Translate (23.77), indicating production-ready quality requiring minimal post-editing. Even languages with modest BLEU improvements show substantial TER reductions, suggesting improved fluency and coherence that traditional metrics may not fully capture. This pattern holds particular significance for scenarios where post-editing cost determines practical viability.

\noindent\textbf{Cross-Metric Correlations.} While surface metrics (BLEU, ChrF++) show high correlation, the divergence between these and neural metrics (COMET, AfriCOMET) reveals important quality dimensions. Languages like Ewe show minimal COMET improvement (0.04) despite substantial ChrF++ gains (30.43 points), suggesting character-level improvements without the corresponding semantic enhancement. In contrast, Arabic shows strong COMET gains (0.33) with modest improvement in ChrF++, indicating semantic preservation despite surface-level challenges. These patterns underscore the importance of multi-metric evaluation for morphologically diverse African languages.

\section{Conclusion}
We have presented the African Languages Lab, a research initiative addressing the critical underrepresentation of African languages in NLP through systematic data collection, model development, and capacity building. Our contributions include a validated dataset of 19 billion tokens and 12,628 hours of aligned speech across 40 languages, substantial performance improvements averaging +23.69 ChrF++ over baseline models, the All Voices platform, and a structured mentorship program developing fifteen early-career researchers. Thus, we demonstrate that the technological marginalization of African languages, while severe, is not intractable.

As language technologies increasingly mediate access to information, education, and economic opportunities, ensuring equitable coverage becomes not merely a technical challenge but a moral imperative. The African Languages Lab demonstrates that this imperative can be met through coordinated research, community engagement, and sustained investment in both technical infrastructure and human capacity, establishing a sustainable path forward for the world's underserved linguistic communities.

\section{Limitations}
\subsection{Model Architecture and Scale Constraints}
Our experiments utilize Llama-3.2-1B as the sole base model, which, while demonstrating the utility of our dataset, may underestimate potential gains achievable with larger-scale architectures. The performance variance across language families, from 63.27 ChrF++ improvement for Swahili to minimal gains for Fon (-1.10), suggests that optimal model selection likely varies by linguistic typology. Additionally, our evaluation of 31 of 40 collected languages reflects FLORES-200 coverage limitations, potentially obscuring insights from the most critically under-resourced languages in our dataset.

\subsection{Dataset Imbalance and Coverage}
Despite assembling 19 billion tokens, our dataset exhibits a 147,000× disparity between the highest-resourced (Amharic: 2,944.95M tokens) and lowest-resourced (Fang: 0.02M tokens) languages. This imbalance directly correlates with performance outcomes: languages with >1B tokens achieve average ChrF++ scores of 45.66, while those with <100M tokens average 24.31. Furthermore, 13 languages lack audio data entirely, limiting multimodal model development. Our validation pipeline, while statistically grounded, operates without native speaker verification for 73\% of languages, potentially missing dialectal variations that affect 28\% of evaluated translations showing COMET-ChrF++ divergence exceeding 0.3.

\subsection{Platform and Infrastructure}
The All Voices platform, while innovative, currently operates primarily through mobile interfaces, which may limit participation from communities with different technology preferences or access patterns. The platform's quality control mechanisms, while systematic, may inadvertently favor certain linguistic varieties over others.

Overall, these constraints delineate several clear pathways for advancement. One direction is to explore architecture-specific optimizations for morphologically complex languages. Another is to implement active learning strategies that help address data imbalances. It is also important to develop evaluation metrics that are more sensitive to African language typologies. These limitations inform our ongoing work and highlight key areas for future research in African NLP.

They also underscore the need for continued investment in computational resources, human expertise, and infrastructure development to support comprehensive technology development for African languages.

\section{Ethics Statement and Broader Impacts}
Developing NLP technologies for LRLs demands rigorous ethical engagement, particularly in contexts shaped by historical exclusion, infrastructural inequities, and linguistic marginalization. Our work at the All Lab is grounded in a principled commitment to the public good, community accountability, and equitable technological development. 

\subsection{Data Collection Ethics}
Our data collection through the All Voices platform operates on principles of voluntary, consent-based participation with full revocability rights. Contributors are informed of their rights, with explicit consent obtained for research purposes including dataset creation, model training, and open-source distribution. The platform implements embedded reporting mechanisms for flagging offensive or culturally inappropriate content, with trained moderators reviewing submissions to maintain quality and cultural sensitivity. We acknowledge that automated validation procedures may miss dialectal nuances or culturally specific meanings, necessitating our ongoing collaboration with native speakers and community experts to expand linguistic coverage and cultural sensitivity.
\subsection{Data Governance.}
Our dataset management follows principles of responsible data stewardship. While we aim for maximum openness, certain data components are subject to agreements with contributing communities that restrict fully public release. We maintain a managed access framework that provides dataset access to qualified researchers while respecting community rights and contributor agreements. Access requests are evaluated based on research purpose, institutional affiliation, and commitment to ethical use.

\subsection{Capacity Building and Research Development}
Our structured research development program has mentored fifteen early-career researchers across four institutions through one-on-one mentorship, project development support, and transitions into extended research roles. This investment in local research leadership establishes sustainable capacity for African NLP development, ensuring that technical advancement aligns with cultural and linguistic expertise. By prioritizing skill development alongside technical innovation, we contribute to a sustainable talent pipeline that positions African researchers to lead future developments in their languages.

\subsection{Societal Impact and Sustainability}
Our work directly advances United Nations Sustainable Development Goals in education and inequality reduction through increased digital representation of marginalized languages. The platform enables community-led content creation and facilitates open knowledge transfer, democratizing access to digital tools while preserving linguistic and cultural heritage. With 88\% of African languages severely underrepresented or completely ignored in computational linguistics, and 814 languages facing extinction risk, our framework provides critical infrastructure for language preservation.

\subsection{Philosophical Framework and Future Vision}
Guided by Ubuntu philosophy—emphasizing inclusivity, interdependence, and openness—we establish a framework for equitable NLP development. Our roadmap encompasses expanding language coverage, optimizing model architectures for low-resource languages, and deepening research collaborations. We acknowledge persistent challenges including limited commercial viability for some LRL technologies and infrastructural constraints, yet our results demonstrate that systematic community engagement can effectively address technological marginalization.
Through this comprehensive approach integrating technical innovation, cultural preservation, educational empowerment, and economic inclusion, we provide replicable models for equitable language technology development that can benefit millions of African language speakers while contributing to global linguistic diversity.

\nocite{ADDA20168, emezue2023afrodigitscommunitydrivenspokendigit,nekoto-etal-2020-participatory,andrew-caines-2019,amol2024statenlpkenyasurvey,10.1145/3567592,elkishky_ccaligned_2020,el-kishky-etal-2021-xlent,banon-etal-2020-paracrawl,elkishky_ccaligned_2020,wenzek2019ccnetextractinghighquality,nllbteam2022languageleftbehindscaling,orife2020masakhanemachinetranslation,joshi-etal-2020-state,magueresse2020lowresourcelanguagesreviewpast}

\clearpage
\bibliography{custom}

\appendix

\section{Aggregated Paper Count Approach}
\begin{table}[H]
  \centering
  \begin{tabular}{l|r|r|r}
    \hline
    \textbf{Source} & \textbf{High-Resource} & \textbf{African} & \textbf{Ratio} \\
    \hline
    GS
    \href{https://scholar.google.ca/scholar?as_q=%E2%80%9Cmultilingual%E2%80%9D+%E2%80%9CSwahili%E2%80%9D+%E2%80%9Clarge+language+models%E2%80%9D&as_epq=&as_oq=&as_eq=&as_occt=any&as_sauthors=&as_publication=&as_ylo=2020&as_yhi=2024&hl=en&as_sdt=0%2C5}{(Link)}
    & 42,871 & 2,121 & 20.2 \\
    arXiv
    \href{https://arxiv.org/search/advanced?advanced=&terms-0-operator=AND&terms-0-term=%22multilingual%22+%22Swahili%22+%22large+language+models%22&terms-0-field=all&classification-physics_archives=all&classification-include_cross_list=include&date-year=&date-filter_by=date_range&date-from_date=2020&date-to_date=2024&date-date_type=submitted_date&abstracts=show&size=50&order=-announced_date_first}{(Link)}
    & 539    & 16    & 33.7 \\
    IEEE 
    \href{https://ieeexplore.ieee.org/search/searchresult.jsp?queryText=%E2%80%9Cmultilingual%E2%80%9D%20%E2%80%9CSwahili%E2%80%9D%20%E2%80%9Clarge%20language%20models%E2%80%9D&highlight=true&returnFacets=ALL&returnType=SEARCH&matchPubs=true&ranges=2020_2024_Year}{(Link)}& 487    & 7     & 69.6 \\
    CORE
    \href{https://core.ac.uk/search/?q=%22multilingual%22+%22Swahili%22+%22large+language+models%22+AND+%28yearPublished%3E%3D2020+AND+yearPublished%3C%3D2024%29&page=1}{(Link)}      & 9,011  & 401   & 22.5 \\
    \hline
  \end{tabular}
  \caption{Aggregate paper counts and ratios between high-resource and African languages (2020-2024). The ratio shows the disparity in research visibility, with higher numbers indicating greater inequality in representation. Search term: “multilingual” “X” “large language models”}
  \label{tab:language-ratios}
\end{table}

\section{Abstract Translations}
\label{app:translations}

\begin{figure}[ht]
    \centering
    \includegraphics[width=\linewidth, trim=0cm 5cm 0cm 0cm, clip]{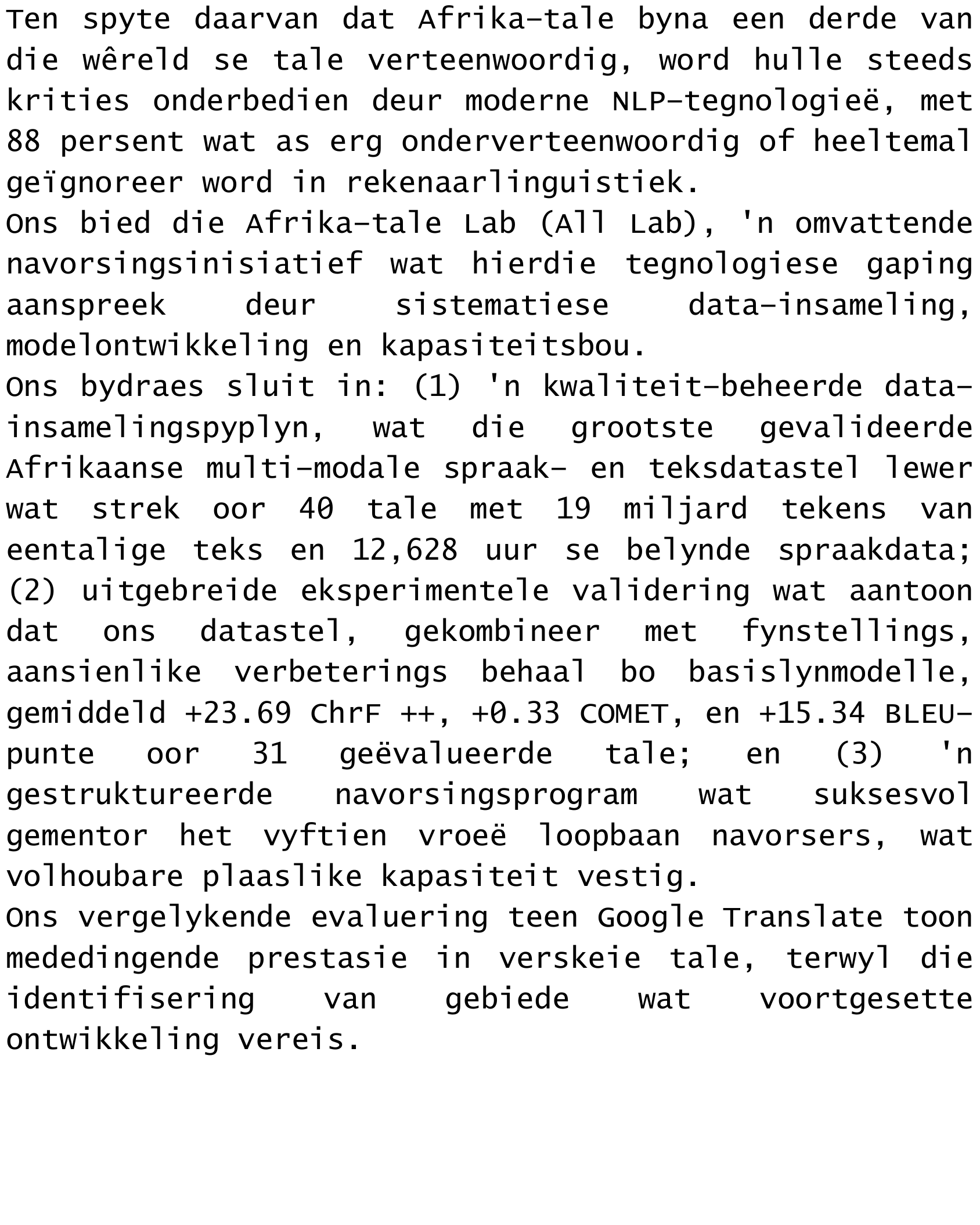}
    \caption{Afrikaans Translation}
\end{figure}

\begin{figure}[ht]
    \centering
    \includegraphics[width=\linewidth, trim=0cm 16cm 0cm 0cm, clip]{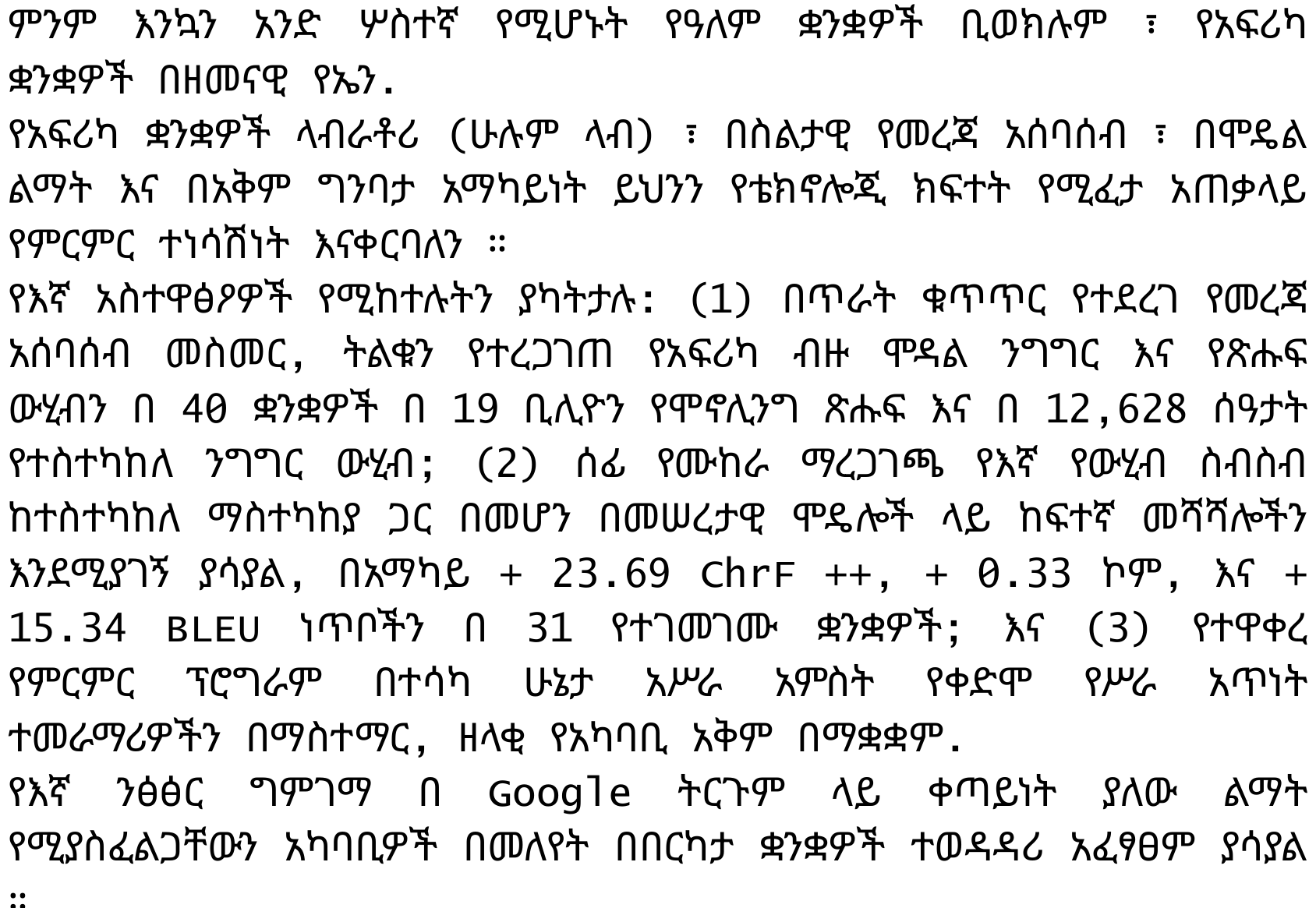}
    \caption{Amharic Translation}
\end{figure}

\begin{figure}[ht]
    \centering
    \includegraphics[width=\linewidth, trim=0cm 1cm 0cm 0cm, clip]{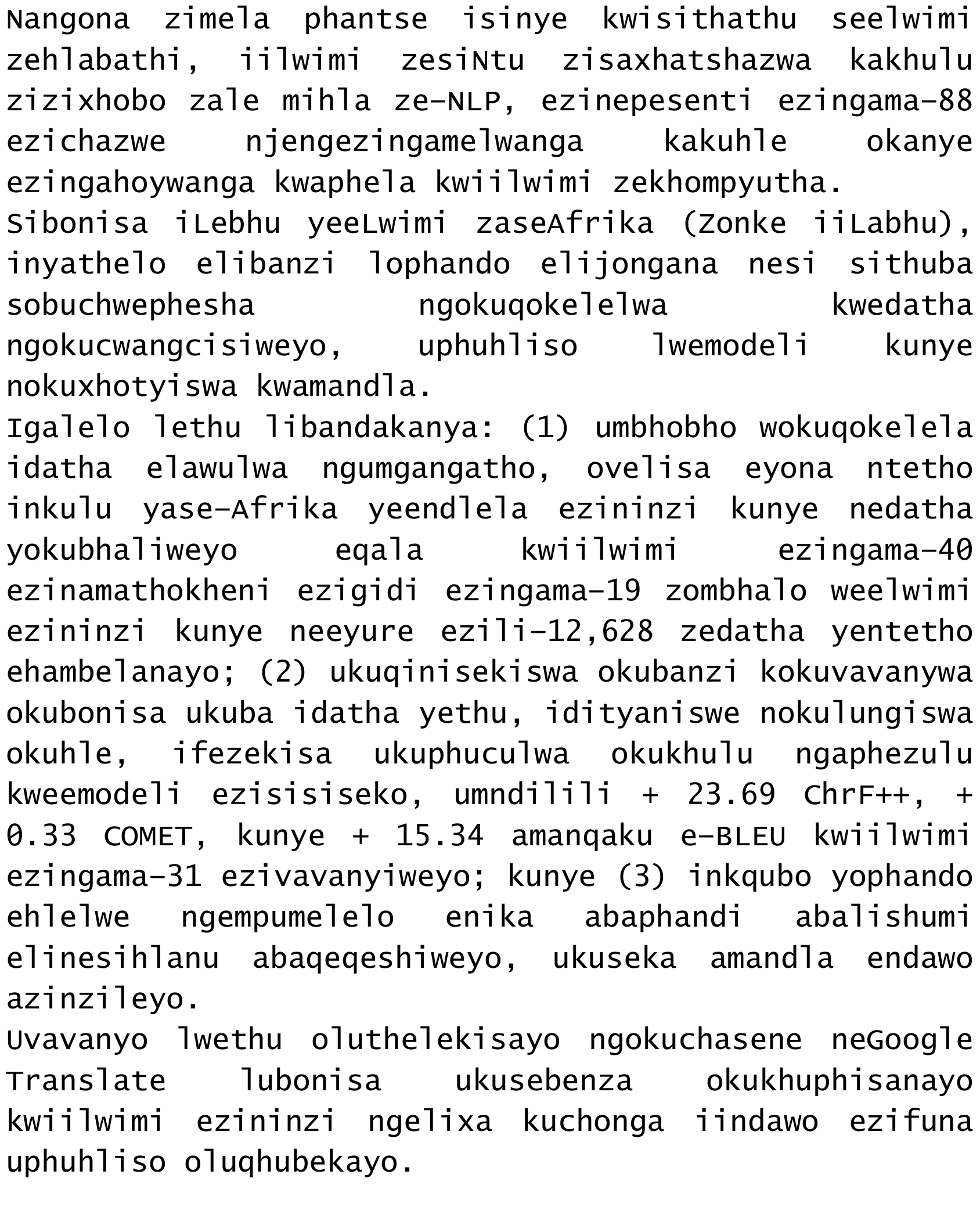}
    \caption{Xhosa Translation}
\end{figure}

\begin{figure}[ht]
    \centering
    \includegraphics[width=\linewidth, trim=0cm 3cm 0cm 0cm, clip]{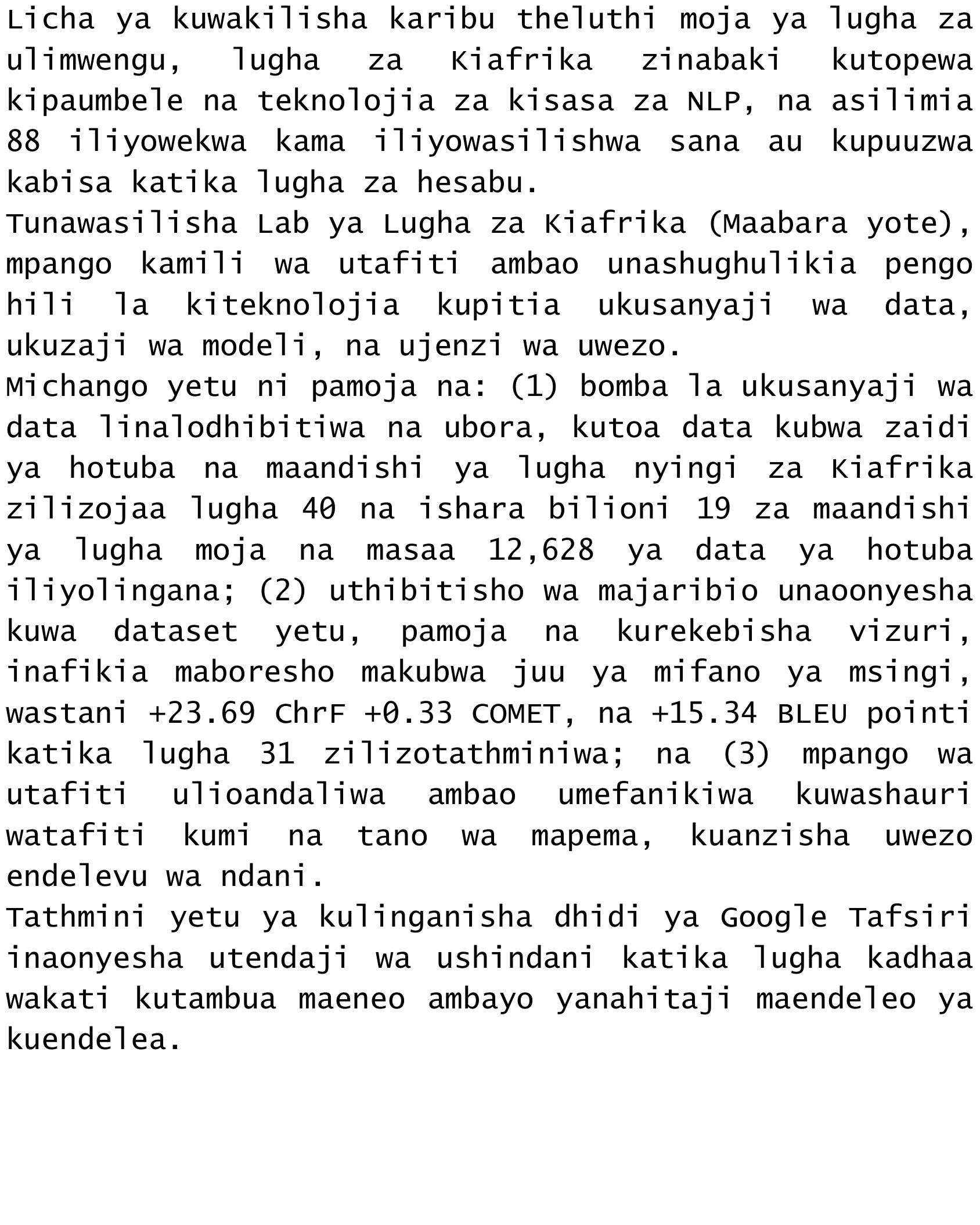}
    \caption{Swahili Translation}
\end{figure}

\begin{figure}[ht]
    \centering
    \includegraphics[width=\linewidth, trim=0cm 4cm 0cm 0cm, clip]{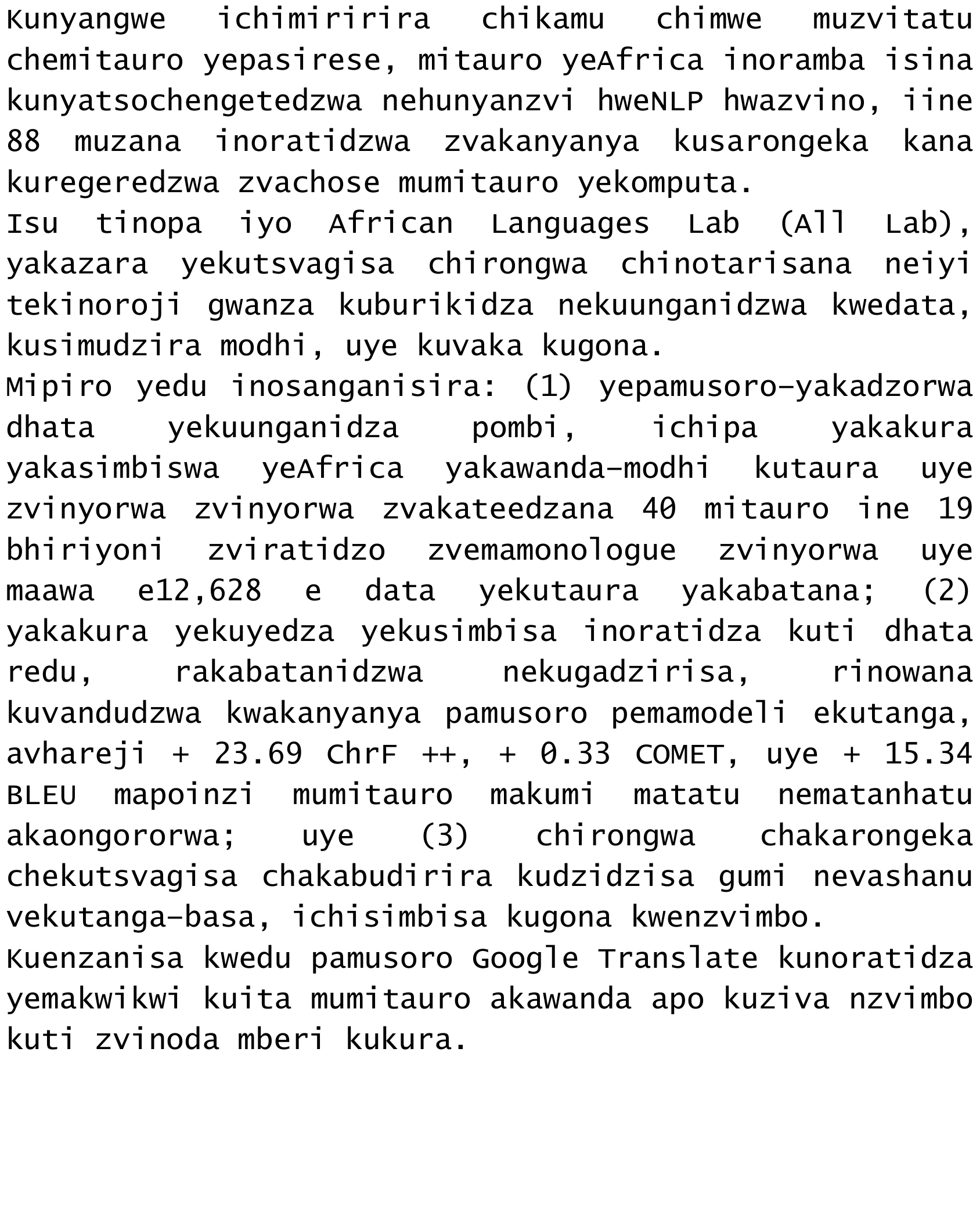}
    \caption{Shona Translation}
\end{figure}

\begin{figure}[ht]
    \centering
    \includegraphics[width=\linewidth, trim=0cm 1cm 0cm 0cm, clip]{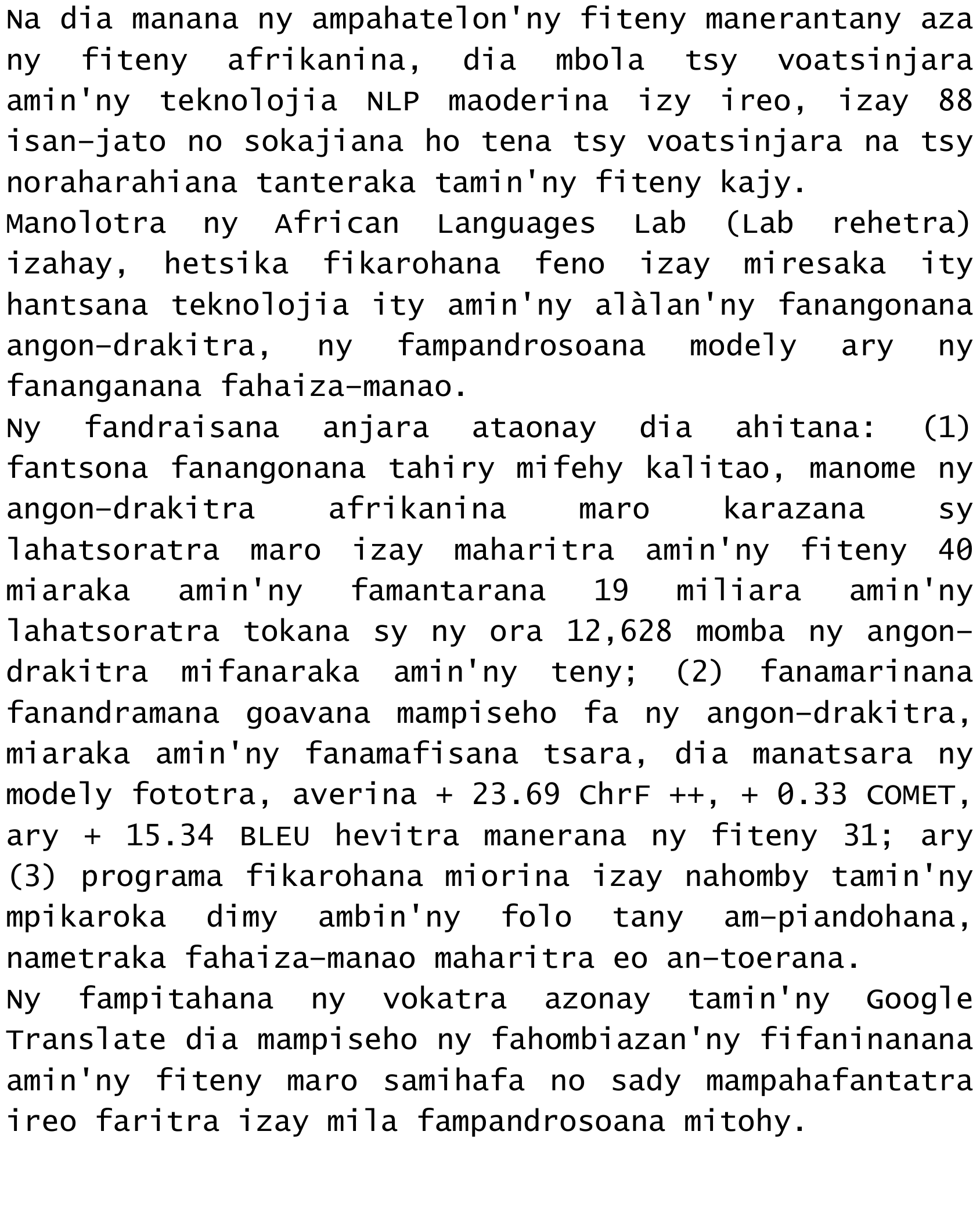}
    \caption{Malagasy Translation}
\end{figure}

\begin{figure}[ht]
    \centering
    \includegraphics[width=\linewidth, trim=0cm 9cm 0cm 0cm, clip]{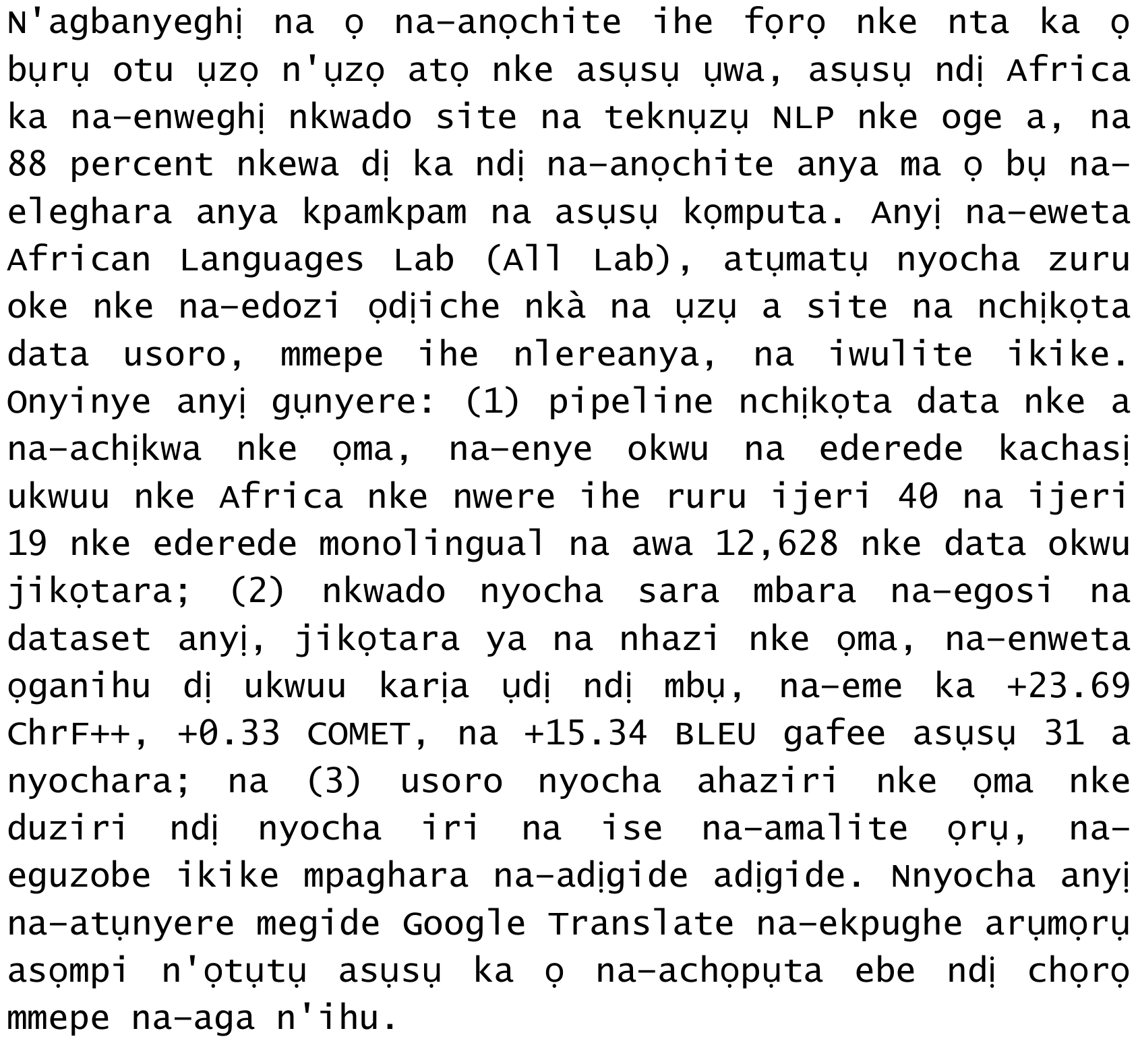}
    \caption{Igbo Translation}
\end{figure}

\begin{figure}[ht]
    \centering
    \includegraphics[width=\linewidth, trim=0cm 6cm 0cm 0cm, clip]{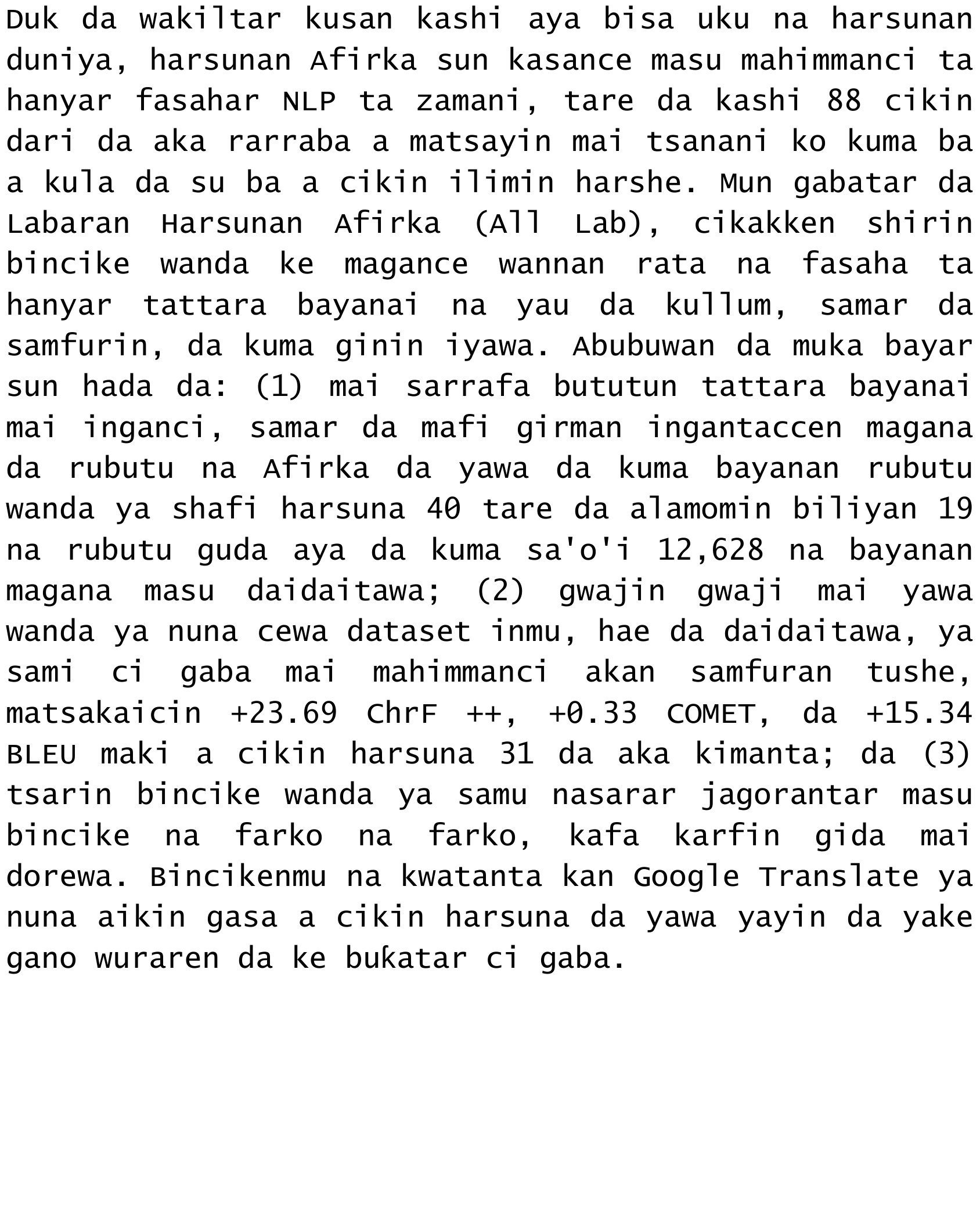}
    \caption{Hausa Translation}
\end{figure}

\begin{figure}[ht]
    \centering
    \includegraphics[width=\linewidth, trim=0cm 15cm 0cm 0cm, clip]{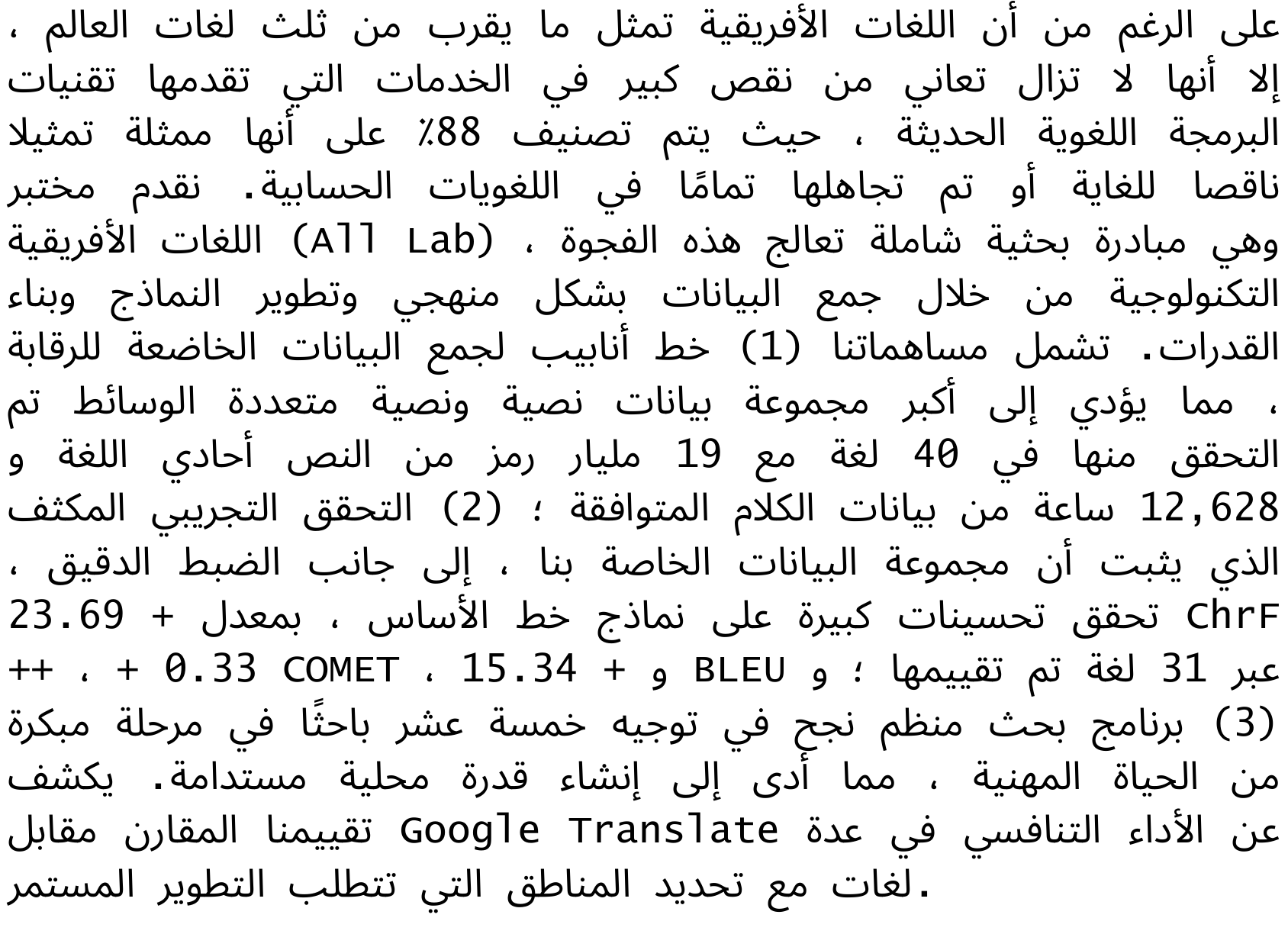}
    \caption{Arabic Translation}
\end{figure}

\begin{figure}[ht]
    \centering
    \includegraphics[width=\linewidth, trim=0cm 1cm 0cm 0cm, clip]{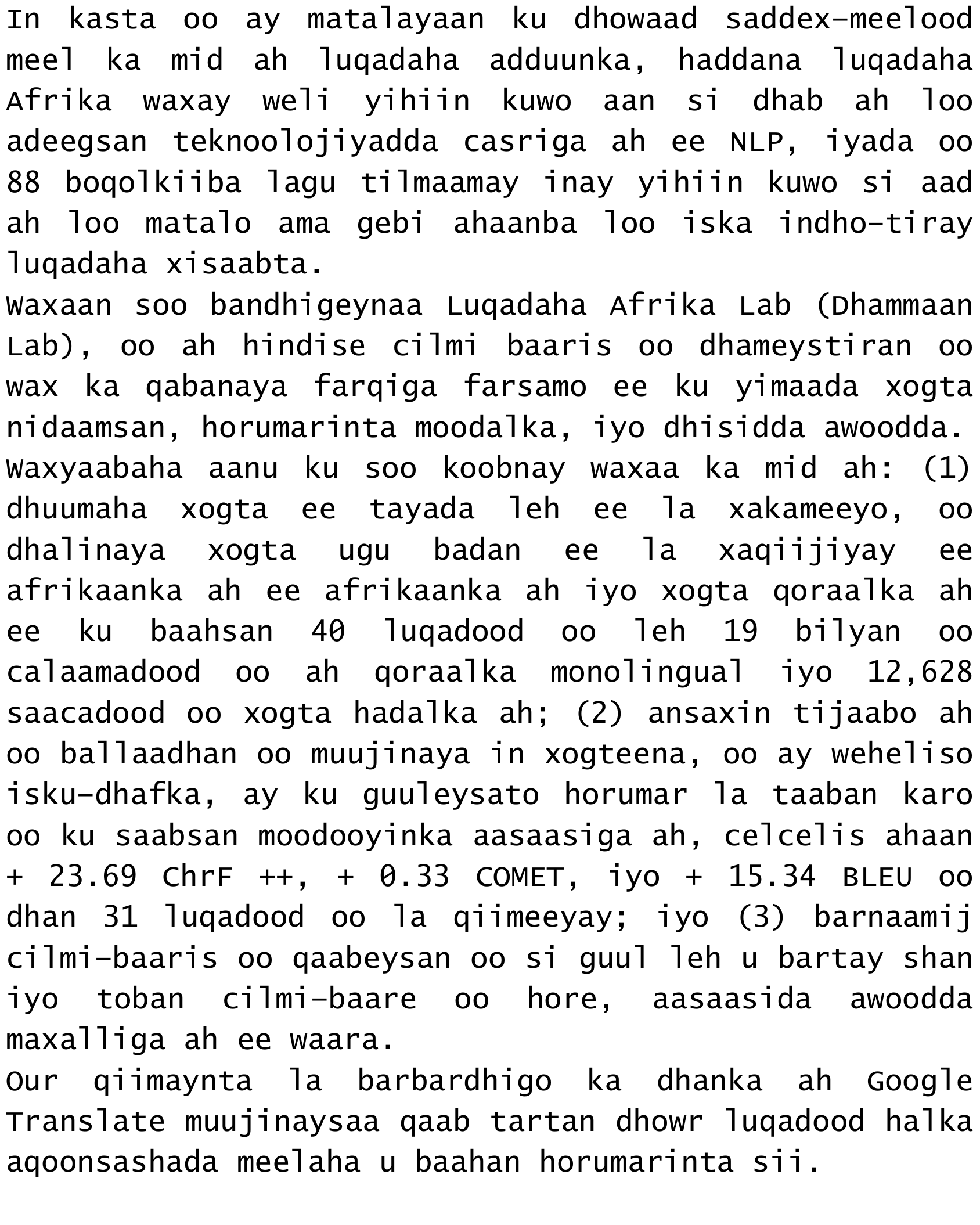}
    \caption{Somali Translation}
\end{figure}

\end{document}